\documentclass[11pt,letterpaper]{article}
\usepackage{times} 
\usepackage{graphicx}
\usepackage{caption,subcaption}
\usepackage[margin=1in]{geometry}
\usepackage[usenames, dvipsnames]{color}
\usepackage{xcolor}
\usepackage{gb4e} 
\usepackage{natbib}
\usepackage{amsmath,amssymb}
\usepackage{multicol} 
\usepackage{multirow} 
\usepackage{setspace}
\newenvironment{nalign}{
    \begin{equation}
    \begin{aligned}
}{
    \end{aligned}
    \end{equation}
    \ignorespacesafterend
}

\begin{document}




       
       
            


\begin{center}
    {\LARGE \textbf{Review of Unsupervised POS Tagging and Its Implications on Language Acquisition}} \\
    \vspace{0.5cm}
    {\large \textbf{Niels Dickson}} \\
    {\large Department of Language Science, University of California Irvine}
\end{center}
\vspace{0.5cm}

\begin{abstract}
    An ability that underlies human syntactic knowledge is determining which words can appear in the similar structures (i.e. grouping words by their syntactic categories). These groupings enable humans to combine structures in order to communicate complex meanings. A foundational question is how do children acquire this ability underlying syntactic knowledge. In exploring this process, we will review various engineering approaches whose goal is similar to that of a child's- without prior syntactic knowledge, correctly identify the parts of speech (POS) of the words in a sample of text. In reviewing these unsupervised tagging efforts, we will discuss common themes that support the advances in the models and their relevance for language acquisition. For example, we discuss how each model judges success (evaluation metrics), the "additional information" that constrains the POS learning (such as orthographic information), and the context used to determine POS (only previous word, words before and after the target, etc). The identified themes pave the way for future investigations into the cognitive processes that underpin the acquisition of syntactic categories and provide a useful layout of current state of the art unsupervised POS tagging models.
\end{abstract}
\vspace{0.3cm}

\section{Introduction}
A foundational goal for the work of language acquisition research is identifying the learning goals children must achieve and uncovering the computations that contribute to their success while learning language. The learning goal investigated in this work is syntactic category induction.

Syntactic categories, groupings of words like nouns or verbs, are essential in understanding the syntactic structure of a language, and learning these groupings allows a child to generalize their knowledge of language to novel instances. For example, understanding that the words ``cat" and ``penguin" belong to the same category allows a learner to generalize the word ``penguin" to novel contexts where ``cat" appears: if one can say ``the cat eats" then ``the penguin eats" may also be a part of the language. Fundamentally, the linguistic principle of productivity is at stake in syntactic category learning. Productivity is the principle of creativity and generalizability in human language where speakers can employ their linguistic knowledge to produce and understand both words in novel contexts and sentences that have never been uttered. 
Illustrating the power of productivity in language, Wilhelm von Humboldt famously remarked that language makes "infinite use of finite means" (\citealt{chomsky1965} quoting \citealt{von1836verschiedenheit}). Acquiring syntactic categories is one of the first steps in leveraging the incredible expressiveness in the structure of human language as these groupings are necessary for building novel structures that follow the rules of the language. 

In what follows, we will begin by more precisely defining syntactic categories which will help motivate why children might care about them 
and how we know that they do care. We will then investigate how children 
could learn these categories by reviewing computational approaches which implement potential mental computations -- ultimately, leading to the popular acquisition model of Frequent Frames \citep{mintz2003}. To better understand the breadth of possible computations that could be applied to this learning problem and in particular, the ones children might apply, we will then review some engineering approaches to unsupervised part of speech tagging. This engineering task is similar to the problem children are solving during syntactic category induction because the engineering models are trying to infer structural categories text alone. But, we will also accompany this comparison with a discussion of how the research goals of acquisition models for syntactic category induction differs from the goals of part of speech tagging. In our discussion of these various computational approaches from the acquisition modeling side and the engineering side, we will focus on four aspects that will guide the synthesis of these approaches in the final discussion. First, what evaluation metrics are they using to judge the success of their model? Second, what additional information are they providing the learner that extends learning past the distributional information present in the text? Third, from a child acquisition standpoint, is the model cognitively plausible? Finally, when labeling a word, what context is the learner paying attention to (some number of words preceding and following a target word or some context of labels)? After reviewing the sample of acquisition and engineering models, we will conclude with a discussion of these four themes to illuminate future directions for research investigating the computations underlying syntactic category learning.

\section{Syntactic Categories}
An important observation about human language is the fact that words can be categorized based on the structures that they appear in. By knowing the structural category, or syntactic category, of a word in a sentence, we can understand how the word fits into the rest of the sentence structure, and ultimately we can understand how the word contributes to the meaning. Despite the foundational nature of these syntactic categories, linguists often struggle to precisely define these categories and to agree on the granularity of the groupings.\footnote{For example, many traditional labels are present in English, but mapping them to other languages may not be straightforward (e.g. it has been argued that Korean expresses the English notion of adjectives with inflected verbs as mentioned in \citealt{jurafsky2009speech}). Further, with what granularity should we group words when many labels have clearly defined sub-classes (e.g. from \citealt{jurafsky2009speech} we see the class of adverbs can consist of directional adverbs, degree adverbs, manner adverbs, and temporal adjectives; from \citealt{syrett2004shifting} we see that children are sensitive to different sub-classes of adjectives, relative, absolute with a maximal standard, and absolute with a minimal standard).} Attempting to define syntactic categories will give us a better understanding of why and how children might learn them.

Much work on syntactic categories relies on our intuitive understanding of categories like nouns and verbs and avoids precisely defining what these groupings are. But as a first pass, let's start with a linguistics textbook definition: Words can be grouped into ``classes called syntactic categories or parts of speech. This classification reflects a variety of factors, including the types of meaning that words express, the types of affixes that they take, and the types of structures in which they can occur" \citep{ogrady2017contemporary}. From this definition, we see that syntactic categories may encode different types of information, and we can see the difficulty in pointing at a particular indicator for identifying these categories. 

One aspect of this definition that seems to ring true for any conceptualization of syntactic categories is that the groupings somehow reflect ``the types of structures in which [the words] occur."
Words that appear in similar structures are a part of the same syntactic categories. This understanding of the structure in language and the groupings that this structure warrants can be traced back to early structural linguists attempting to analyze languages (\citealt{mintz2002distributional} names Leonard Bloomfield and Zellig Harris as early structural linguists). \cite{harris1946} provides an early procedural test for starting this linguistic analysis: ``morpheme classes are formed by placing in one class all morphemes which are substitutable for each other in utterances, as \textit{man} replaces \textit{child} in \textit{The child disappeared}." So, applying this procedure to our earlier example, we know that ``cat" and ``penguin" belong to the same category in English because substituting one for the other in a syntactic sentence produces another grammatical sentence: ``the cat eats" and ``the penguin eats." Despite the difficulties we've encountered in precise definitions for syntactic categories, what seems to be uncontroversial is that these categories are closely tied with language structure. In human language, some words can appear in the same structure while others cannot. This basic observation will help motivate our investigation into the acquisition of syntactic categories.

\section{acquiring syntactic categories} \label{sec:acqgm}
Syntactic categories have become such an important tool for linguistic analysis that the psychological reality of these categories have often been assumed. When transitioning to the investigation of learning these groupings, we need to be careful in our assumptions of what children know and what they are trying to learn. We have motivated the importance of syntactic categories in terms of productivity and generalizability which help answer why a learner might care about these groupings.
But what evidence do we have that children are sensitive to the groupings linguistics have proposed? And how might children learn these groupings.
\subsection{psychological reality of syntactic categories}
To investigate the psychological reality of syntactic categories, we can first get a glimpse into the groupings that children are making through overgeneralization errors they make. For example, when children are acquiring the past tense in English, they often overgeneralize the rule that adding ``-ed" to a verb will produce the past tense. They may produce a verb like ``go-ed" instead of the irregular past form ``went." This overgeneralization can tell us which words the child groups into a verb-like category because the children are systematically applying this rule to a certain group of words. 
More generally, this example further motivates the importance of syntactic categories for a language user because the child 
needs to determine which category 
of words to apply certain affixes to. 


Second, we can investigate children's syntactic groupings through experimental work using nonce stimuli. \cite{shi2010syntactic} found that children as young as 12 months old were forming some syntactic groupings by leveraging function words. The study, conducted with French stimuli, tested French children using a head-turn preference paradigm. The participants were familiarized with a noun category context for a nonsense word (e.g. ``some blicks" and ``your blick"). Then, during testing, the children were able to differentiate, evidenced by significantly different looking time, between a novel noun context and an ``ungrammatical" verb context (e.g. ``the blick" versus ``you blick"). 
This result suggests that children are able to categorize novel words based on limited context.
These pieces of evidence illustrate that children are sensitive to syntactic categories, 
and they are engaging in some structural grouping procedure. 

\subsection{proposals for learning syntactic categories}
In beginning to think \textit{how} they may accomplish this, the formulation of syntactic categories from \cite{harris1946}, using distributional information from structural contexts, provides a productive starting place. \cite{maratsos1980internal} followed this distributional procedure and proposed that children may be relying on distributional information in speech to learn syntactic categories. In particular, the authors suggested that a child could group words that overlap in frequently occurring phrases. While we will largely be focusing on the distributional evidence available to children and following this distributional formulation, it is important to note that this account is not uncontroversial. \cite{mintz2002distributional} outlines two theories that propose that distributional evidence only provides a secondary role  in syntactic category acquisition. The first theory suggests semantic categories form the basis of syntactic categories. 
Of course there are some outlier examples where the syntactic category may not be extractable from the meaning alone which presents some difficulty for this theory (nouns like wiggle or step and abstract words like think, love, or know). But, this approach may still be valuable for the distributional account because we know that children are juggling many learning goals in the early stages of language learning (e.g. speech segmentation, phoneme identification, etc.). So assuming that children learn a word's syntactic category independent of its meaning seems implausible given the role of meaning in children's early language goals. As we will see, the semantics and distributional approaches can meaningfully interact as semantic information may help bootstrap children's syntactic category learning from distributional evidence \citep{gutmanetal2015, stratos2016unsupervised}. \cite{mintz2002distributional} calls the second theory the nativist theory which suggests that the set of possible syntactic categories is innately specified \citep{chomsky1965, pinker1984}. This specification substantially constrains the learning process where the learner still relies on distributional evidence to assign the words to categories, but they do not have to infer which categories exist, and they do not have to consider different sets of categories of varying lengths. In the interest of minimizing the information that we assume children must bring to the learning problem, we will explore how much children may be able to learn from distributional information, and we will see how additional information available to children in their language input could further aid this process. 
In the following section, and for the bulk of this paper, we will turn towards that potential computations involved in children's syntactic category groupings to get a better understanding of how children perform these groupings.

\section{potential computations in learning} \label{sec:complearning}
Following the proposal from \cite{maratsos1980internal} suggesting that children may learn syntactic categories from distributional evidence, \cite{cartwright1997syntactic} attempted to computationally implement the strategy of grouping words based on overlap in frequent phrases. \cite{cartwright1997syntactic} attempted to identify ``minimal pairs" in child directed speech (CDS) where two sentences only differed by one word (e.g. ``My cat \textit{meowed}" and ``My cat \textit{slept}" identifies a grouping of \textit{meowed} and \textit{slept}). This work did illustrate the plausibility of learning groupings similar to syntactic categories from distributional information, but this method is computationally intensive and not plausible for a children's learning algorithm. 
Another computational implementation that leveraged distributional information was \cite{redington1998distributional}. The authors chose the 1000 most frequent words in the corpus as target words to categorize. They then leveraged distributional information in the form of context vectors (a window of two words preceding and two words following the target word. 
They then calculated similarity between the vectors and iteratively clustered the most similar words. Again, this work supported the proposal that groupings similar to syntactic categories could be recovered from distributional information, but this method was also computationally intensive and assumed an implausible memory capacity for child learners. Together, these efforts provided evidence for the usefulness of distributional information, but the practicality of children implementing these algorithms remained uncertain.  

These works laid the foundation for leveraging distributional information during syntactic category learning, but building off these efforts, \cite{mintz2003} proposes a novel categorization approach which has greatly shaped how language acquisition researchers conceptualize category learning \citep{wangmintz2008, chemlaetal2009, wangetal2011, moran2018universal}. The authors introduce the Frequent Frame (FF) which uses the frequent nonadjacent words in a corpus to group words into syntactic categories.
A frame consists of three words, A\_B\_C, where B is the ``target word." By locating all the instances of the A\_X\_C frame, where X is allowed to vary, we can group all the X's that appear in this context. We get the label of ``frequent" for these frames because of all the frames that appear in the corpus, we select the most frequently occurring frames with the assumption that these will be most informative for extracting distributional information (in the original formulation, \citealt{mintz2003} rather arbitrarily selects the 45 most frequent frames, but later work grapples with determining the frequency cutoff, as in \citealt{chemlaetal2009} where they present FF performance by iteratively including less and less frequent frames as opposed to reporting performance for only the most X frequent frames).

In contrast to the past learning attempts from distributional information, the FF learning procedure is computationally simple. While \cite{cartwright1997syntactic} assumed the child categorized words in the encountered sentence and then merged with previous encountered frames based on certain criterion and \cite{redington1998distributional} assumed the child represents a context vector of frequent words for the target words and then iteratively grouped based on similarity score, \cite{mintz2003} only assumes that the child is counting these frames and then grouping the words seen in the frequent ones. 
Further, without assuming the implausible two-step procedure of counting the frames in the corpus to identify the frequent ones and then listing all the intervening words in the frame, \cite{wangmintz2008} showed that a more plausible one-step FF categorization performed well. The authors showed that iteratively exposing the language input to a model with a limited memory of the encountered frames can produce similar FF results. 

As we have seen, the FF algorithm is quite simple. This simplicity can partially explain the algorithms popularity in the acquisition literature because we can be confident in the cognitive plausibility of simpler algorithms. But other advantages of the FF approach help explain its success. 
In its initial formulation tested on English CDS data, FF produced very accurate clusters for frequently occurring syntactic categories \citep{mintz2003}.
Another advantage of FFs comes from evidence that young children ($\sim$ 12 months old) are sensitive to these frequent frames when encountering novel words which suggests that they can use FF contexts to group words \citep{mintz2006}.  

Despite all of its promise, there still remains some fundamental questions around syntactic category learning with FFs. I mentioned that evidence from English suggests that FF clusters are very accurate for frequent syntactic categories, but FF learning offers very limited coverage of the words in the tested corpus.
The accuracy measures the number of hits, or true positives, (noun properly in a noun category) while taking into account the false positives (an adjective incorrectly in a noun category). On the other hand, completeness measures the coverage of the categorizing technique which measures the number of true positives while taking into account false negatives (a noun which was not labelled as a noun). The groupings from FF have notoriously poor completeness as the frames create many small clusters of nouns or verbs for example but limited larger clusterings resembling the hand tagged groupings from the corpus data. These initial results of high accuracy but low completeness has shaped the formulation of FF as researchers proposed that this categorization technique may allow learners to form accurate initial groupings and then reassess and potentially merge clusters when necessary for the learner.
\begin{equation} 
    accuracy = \frac{true\: positives}{true\: positives + false\: positives}
\end{equation}
\begin{equation} 
    completeness = \frac{true\: positives}{true\: positives + false\: negatives}
\end{equation}

Perhaps an even more foundational question to the FF approach deals with crosslinguistic data and the level of linguistic analysis 
over which FF should operate. Attempts to extend the English word level FF results to other languages has produced mixed results (success in French, \citealt{chemlaetal2009}, and German, \citealt{wangetal2011}, but poor performance in more morphologically complex languages like Turkish, \citealt{wangetal2011}. See \citealt{moran2018universal} for a summary table of cross linguistic data). This competing cross linguistic evidence raises the questions of whether the relevant unit may differ across languages (\citealt{moran2018universal} suggest that the morpheme level is more appropriate for some languages) and how children can identify the relevant unit. 


Considering these outstanding questions in conjunction with the popularity and success of the the theory of FF, a review of the computational proposals for learning syntactic categories would help us situate FF and the acquisition modeling approach within the potential computational architectures and better understand the dimensions that could vary in our computational theories.
To this end, we will review the engineering approach to part of speech (pos) tagging to inform our understanding of a child's computations in syntactic category learning. The engineering side offers a wide range of model architectures (incorporating various learning algorithms and learning from information beyond distributional inforamtion) that are not often considered by the acquisition modeling side,
so this crossover has the potential to inform our acquisition models of syntactic category learning in addition to exploring the multiple cues a learner could leverage from their input. 

\subsection{acquisition models inspired by engineering approach}
Before turning to the engineering models, I want to present the remaining acquisition models we will investigate which will highlight how the crossover between the engineering models and the acquisition models has been a fruitful research area for investigating learning of syntactic categories. Many acquisition researchers have leveraged the more technical engineering models to develop models that attempt to illuminate children's syntactic category learning and extend upon the simple distributional learning with FF. The dominant model framework from the engineering side which will be used in the next acquisition model is called a Hidden Markov Model (HMM). More details of the HMM will be presented in section \ref{sec:van}, but figure \ref{fig:hmmgraph} illustrates the general structure where the words for some language data are the observed data points, $w$, and we attempt to learn the hidden states, $y$, corresponding to syntactic category labels which we assume has some hand in generating the words in the data. For an HMM, the relevant parameters that must be learned are the transition probabilities, the probability of producing a subsequent tag from the current tag (i.e. for a tag, $t_i$, we learn the probability of transitioning to any other tag), and the emission probabilities, the probability of emitting a certain word from a tag state. One of the main intuitions from the engineering work is that the basic HMM structure may be inappropriate for syntactic category learning,
so, as we will see, much of the work attempts to extend the basic architecture to incorporate additional information beyond distributional information. In a similar fashion, we can see how researchers interested in child language learning have proposed additional information that children may learn from during syntactic category learning.


One additional piece of information that may aid the child learning process is sentence types. \cite{franketal2013cat} use the sentence types labeling of declaratives, yes-no questions, wh-questions, exclamatory, and imperative to augment a model of syntactic category learning. The authors propose that different sentence types will have different syntactic distributions, so knowing the sentence type will help inform our expectations of the syntactic context. For example, the distribution of syntactic categories in imperatives (e.g. ``Pick up the toy") will be different than the distribution for a simple declarative as you are more likely to see a verb as the first word in the imperative sentence. They also motivate the acquisition claim that children have access to sentence type information while learning syntactic categories by citing evidence that prosody strongly predicts sentence types in most languages and children are sensitive to prosodic cues early in development \citep{hirst1998survey, homae2006right}. Additionally, the authors note that the distribution of sentence types in CDS differs from the distribution in typical adult directed speech as the sentence types in CDS are much more diverse \citep{fernald1991prosody}. Showing that children could leverage sentence type information during syntactic category learning would motivate this feature of CDS. 

To test whether adding sentence type information may be helpful to a child learning syntactic categories, \cite{franketal2013cat} begin with a Bayesian HMM model \citep{goldwatergriffiths2007} as their baseline which follows the same structure as the distributions defined in section \ref{sec:extvan} with equation \ref{BayHMM}. 
We see from these distributions that a transition probability depends on a parameter $\theta_y$ where $y$ is the previous tag. In order to extend this model definition to include sentence types, \cite{franketal2013cat} change the parameterization to include a variable for sentence type $s$ such that a transition probability now depends on a parameter $\theta_{y,s}$. Therefore, the model must learn the transition probabilities for each tag and for each sentence type. As alluded to earlier, this formulation corresponds to the intuition that the tag distributions are affected by the sentence type such that transition probabilities may look very different across these types. The authors then tested the model on English CDS from the CHILDES collection of corpora \citep{macwhinney2000}. Testing on CDS produces a more cognitively plausible learning environment for the model, 
and this procedure sets the acquisition modeling approaches apart from the engineering approaches which test exclusively on adult text corpora. The authors found that the model that included sentence types performed equal to or beyond the baseline bayesian HMM for each of the sentence types.\footnote{The authors needed to test the models on data from a specific sentence type separately because the HMM with sentence types is learning different distributions for each type. This means that a particular declarative tag does not match the corresponding imperative tag so evaluating the tagging as one model would be incorrect.} 
Although follow up work using data from other languages produced mixed results on the utility of sentence types, \cite{franketal2013cat} conclude, based on their results overall, that sentence types, a piece of information young children may have access to, can assist in the process of learning syntactic categories.

\cite{gutmanetal2015} proposed another piece of information that may assist in children's syntactic category learning: lexical meaning. Although the authors were not specifically looking at labeling words, \cite{gutmanetal2015} investigated a related task of how children label prosodic phrases with syntactic labels (e.g. noun phrase, verb phrase, etc.).
Again, the authors motivated their computational approach with evidence from acquisition research in order to illuminate the learning process for children. They motivate this investigation into prosodic phrase categorization by citing evidence for children's sensitivity to prosody from birth \citep{mehler1988precursor} and referring back to past proposals that prosodic phrases may play a role in bootstrapping the acquisition of syntax \citep{morgan1986}.

To model children's prosodic phrase labeling, \cite{gutmanetal2015} implement a Naive Bayes model to infer a label for each phrase using some predictor variables (corresponding to the words in the phrase). For this work, the authors used French CDS that has been segmented by prosodic phrases (Lyon corpus in the CHILDES database: \citealt{macwhinney2000}). From this input, the model first initializes the groupings using the first word in each phrase (as, in French, these will often be frequent, content words that may assist in categorization), then relabels with a type of EM learning that maximizes the probability of the predictor variables for a phrase given the selected label. Intuitively, we want the labeling scheme that best predicts the words in the phrases. Evaluating their model based on cluster purity,\footnote{To calculate cluster purity, the authors first matched the phrases to their label in the gold standard grouping. Then for each cluster, you identify the most common gold standard label and calculate the proportion of the cluster which follows this label.} the authors found that the model created clusters with relatively high purity suggesting that children may be able to correctly categorize the phrases in their input. But, in analyzing the groupings, they found that the model does not create groupings corresponding to traditional noun phrase, verb phrase, etc. labeling, but instead, these labels are distributed across many smaller model labels. This problem is reminiscent of the incompleteness problem faced by the FF approach.

To overcome this deficiency, the authors extended their model and assumed another piece of information that a child may leverage to learn these labels. In particular, they assume that children engaged in this learning problem will have already learned a small list of word meanings.
Using these word meanings, children may be able to initialize their prosodic phrase labels where the meaning of a small number of physical objects (toy, car, teddy bear) may form the seed of a noun-like category and a small number of action meanings (eat and play) may form the seed of a verb-like category. Narrowing in on the performance of the model at identifying noun phrases and verb phrases, the authors found that initializing the model with the semantic seed, labeling a phrase noun when we see one of our physical object meanings or verb for action meaning, produced a large increase in accuracy and completeness. For example, even with a small list of assumed word meanings (6 nouns and 2 verbs which were the most frequent in the input), accuracy jumped from $\sim$0.4 (with no lexical seeding) to $\sim$0.75 and completeness jumped from $\sim$0.4 to $\sim$0.6. 
Although this work is modeling a different learning problem than the current focus of syntactic categories, they are concerned with the child's learning process of syntactic phrase groupings. The novel contribution of this work is the implementation of semantic seeding which enables the learner to leverage a small list of word meanings to initialize syntactic groupings. This contribution has its roots in the semantic category hypothesis outlined in Section \ref{sec:acqgm}, and we have seen how this hypothesis has been explored in both the engineering approaches (e.g. the anchoring technique in the A-HMM) and the acquisition modeling. We will return to this connection in section \ref{sec:addinfo}.

\section{engineering side}
With the label ``engineering," we are grouping the part of speech (\textbf{pos}), another term for syntactic categories, 
learning approaches that are concerned with performance of their models with the goal of tagging some text as closely to a hand tagged corpus as possible. 
Identifying the parts of speech in a corpus is useful for a wide variety of analyses in Language Science work, but working through a corpus to label each word as a noun, verb, etc., tagging by hand, is time consuming and expensive. Additionally, hand tagging may be infeasible when attempting to work with larger and larger corpora.
By automating this tagging process, the engineering approach is attempting to overcome these hurdles in achieving tagged linguistic data \citep{stratos2016unsupervised}. 

From the outset, we can see that the goals of the engineering approach diverge slightly from the language acquisition approach. The engineering approach builds models that measures success based on the learning of the gold standard, hand tagged groupings, while the syntactic category learning models attempts to incorporate linguistic cues available to children with computations that children may be performing for a realistic learning of syntactic categories.
To this end, while both may measure performance against gold standard tags, the acquisition modeling side often focuses more attention on frequent content categories 
and the possible ungrammatical groupings from the input 
while the engineering approach is solely focused on achieving the gold-standard labeling.

The engineering pos tagging models can be divided into supervised and unsupervised models.
Here, the supervision refers to whether the model provides a partial dictionary specifying the category, or categories, that a word belongs to. The supervised model provides these categories so the model uses the dictionary to assign some categories for the text input and disambiguates for the words with multiple possible categories using contextual information. On the other hand, the unsupervised models do not have a dictionary to assist the in tagging as they must group words in the testing text data only based on the distributional information. This approach is especially useful when tagging cross-linguistic data where a dictionary may not be available. For this review, we will focus on the unsupervised models as this task more closely resembles a child learner's task because children are learning syntactic categories from the input of speech sounds without being exposed to the concept of syntactic categories (children may be explicitly taught syntactic categories later in school, if at all, after their tacit knowledge has begun developing).


\subsection{model evaluations} \label{sec:modeval}
Before outlining model details for unsupervised pos tagging, we first need to understand how these models will be evaluated and compared in this literature. These models take a corpus of text and assigns a label to each word in the corpus, so the traditional evaluations attempt to measure how closely this labelling aligns with the gold standard tags. A challenge for this process arises from the fact that these unsupervised models do not have knowledge of the gold standard tag list.
So the models are creating tags without direct correspondence to the gold standard list (noun, verb, etc.) which necessitates mapping the model's tags to the gold standard tags in order to evaluate. One popular metric following this procedure is the many-to-one accuracy \citep{christodoulopoulos2010two}. This metric first maps each model cluster to a gold standard tag which is most common in that cluster (e.g. if the most frequent category of the words in a particular cluster is the noun category, then this cluster will be labeled as nouns). This metric is called many-to-one because many clusterings can be mapped to the same gold standard tag. With the words now labelled using a gold standard tag, many-to-one accuracy is calculated as the percentage of correctly labeled words (measure can range from 0 to 1 where 1 would be all words labeled as the gold standard). In a similar way, we can also evaluate the model's performance using a more stringent mapping procedure. This metric is called one-to-one accuracy because each gold standard tag can only correspond to one model cluster. This measure is calculated by first greedily mapping each cluster to a gold standard tag until no tags or no clusters remain, then calculate the accuracy of the labelled corpus based on the gold standard labels (again ranging from 0 to 1). 

While these two may be the most popular accuracy measures, other information-theoretic measures have increased in popularity. In a review of pos tagging evaluation metrics, \cite{christodoulopoulos2010two} note that the popular many-to-one and one-to-one measures are not stable as the number of clusters increases. So, as the number of clusters increase, the many-to-one accuracy tends to increases with the undesirable feature of perfect accuracy with each word receiving a unique label (as a result of mapping many clusters to one tag). Alternatively, as the number of clusters increases, the one-to-one accuracy tends to decrease with the more stringent mapping restriction. This unstable feature of the accuracy makes comparing models with different cluster numbers difficult. To overcome this problem, the information theoretic measures treat the model clusterings and the gold standard tagging as two distributions and quantifies the difference in information content between these two distributions. Additionally, these measures do not require mapping the model clusters to the gold standard tags as the clustering is evaluated as a whole as opposed to making an assumption on which tag a certain cluster corresponds to. These information theoretic measures rely on the measure of conditional entropy, where they can account for how much information is lost going from the model clusters, $C$, to the gold standard tags, $T$, and how much information is lost going from gold standard tags, $T$, to model clusters, $C$. Formally these entropy measures are $H(C|T)$ and $H(T|C)$ and can be interpreted as how much information remains in the distribution of $C$ after observing $T$ and vice versa \citep{christodoulopoulos2010two, meilua2003comparing}. When reviewing the unsupervised pos tagging models, we will often report the many-to-one accuracy due to its popularity, but we will return to evaluation metrics in section \ref{sec:evals}. 

\subsection{Vanilla HMM} \label{sec:van}

\begin{figure} 
    \centering
    \includegraphics[width=0.5\linewidth]{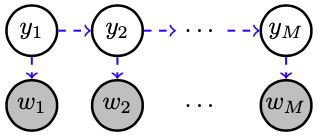}
    \caption{Graphical representation of an HMM taken from \cite{eisenstein2019introduction}.}
\label{fig:hmmgraph}
\end{figure}

Diving into the unsupervised pos tagging literature in the hopes of informing our computational understanding of children's syntactic categorizing, we begin with the most popular model architecture: Hidden Markov Model (HMM). The HMM relies on the intuition that the observed data sequence can be generated 
by unobserved, hidden states. This intuition naturally applies to the pos tagging process where the observed states are the words we observe in corpus and the hidden states are the parts of speech that can account for the particular words that we see. In the traditional HMM structure, illustrated in figure \ref{fig:hmmgraph}, we see to generate some particular observed data, $w_1, w_2, ..., w_M$, the model starts at a particular hidden state, $y_1$, which emits a word, $w_1$, before transitioning to another hidden state, $y_2$, and so on. Importantly, the model relies on a Markov assumption (equation \ref{eq:markov}) such that the probability of a specific word only depends on the hidden state and the previous hidden state as opposed to a larger context window: 
\begin{align} \label{eq:markov}
    \text{p($w_i|$context) = p($w_{i}|y_i$) $\cdot$ p($y_i|y_{i-1}$)} 
\end{align}
With this model structure, an HMM pos tagger must learn the probability of transitioning from one hidden state to another, transition probability, and the probability of emitting an observed state from a hidden state, emission probability. 

Comparing the HMM structure back to the popular acquisition modeling approach, FF, we can start differentiating the FF architecture from other possible structure. With FF, the learner acquires a syntactic category for a certain word depending on the two surrounding words. But, under the HMM structure, a particular word's category label depends on the previous word only through that previous word's category label.
So, the transition tag distribution dictates the particular tag. 

With the basic HMM structure outlined, we will now turn our attention to the learning algorithms that allows a model learner to learn the relevant parameters from the distributional information. The most traditional learning algorithm that is applied to the HMM model structure is a maximum likelihood estimator called Expectation Maximization (EM) which attempts to find the parameters that maximize the log-likelihood of the data 
\citep{johnson2007doesn}. EM converges to a local maximum of objective by first initializing the parameters (i.e. specifying the transition probabilities and emission probabilities),
and then estimating the model counts from this initialization (Expectation step), and finally, using these counts to re-estimate the model parameters (Maximization step). 

One of the main intuitions from the engineering side, the unsupervised pos tagging literature, is that learning an HMM using EM produces relatively poor performance on pos tagging (\citealt{johnson2007doesn} theorized about the cause of this poor performance based on the tag distribution. \citealt{stratos2016unsupervised} provide evidence for its poor performance in comparison with other models). This architecture, learning an HMM using EM, is known as a vanilla HMM, and because of its poor performance 
(around 0.62 many-to-one accuracy reported by \citealt{johnson2007doesn,stratos2016unsupervised}), much of the work in this literature has focused on extending this basic architecture to improve performance. So, we will now turn to these extensions which will help outline the possibilities in the variation in computations underlying syntactic category learning.

\subsection{extensions to vanilla HMM} \label{sec:extvan}
\begin{figure}
    \centering
    \includegraphics[width=0.5\linewidth]{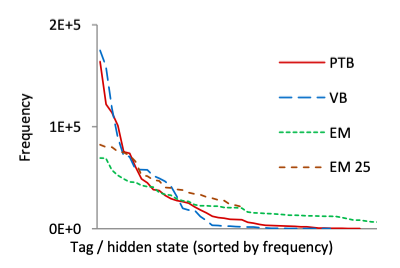}
    \caption{Taken from \cite{johnson2007doesn}, this plot illustrates the number of tokens (y-axis) assigned to each tag (x-axis). We can see the peaked tag distributions from the gold standard tags of Penn Treebank (PTB) and the HMM learned by VB. Contrasted with the flatter tag distributions of the HMM learned with EM and EM-25 (where the number of hidden states is set at 25 as opposed to 45 corresponding to each tag in the PTB).}
\label{fig:JohnsonTagDist}
\end{figure}

The first extension to the Vanilla HMM varies the algorithm used to learn the HMM parameters. Where the vanilla HMM uses the EM learning algorithm, \cite{johnson2007doesn} proposes using a Bayesian learning algorithm. The author noticed that the tag distribution learned with the vanilla model did not match the distribution from the actual gold standard tag distribution. In particular, the gold standard tags have a peaked distribution where a few of the tags are used very frequently while many of the tags are infrequent, but EM learns a much flatter distribution of tag frequencies where the word tokens are more evenly distributed across the different tags (as seen in Figure \ref{fig:JohnsonTagDist}). To alleviate this discrepancy, \cite{johnson2007doesn} proposes a Bayesian learning algorithm where one can control the sparsity of the emission and transition probabilities to manufacture this peaked tag distribution.

To illustrate how the Bayesian learning more closely mirrors the actual tag distribution, we observe how the Bayesian framework extends the model structure. \cite{johnson2007doesn} defines the standard HMM model \eqref{HMM} as distributions over hidden states, $y$, and observed states, $w$, sampled from multinomial distributions. 
\begin{nalign}
\label{HMM}
    &y_i\ |\ y_{i-1}=y\ && \sim\ && \text{Multi}(\theta_y) \\
    &w_i\ |\ y_{i}=y\ && \sim\ && \text{Multi}(\phi_y)
\end{nalign}
The hidden state distributions defines our transition probabilities (where the model needs to learn the $\theta$ parameters) and the observed state distributions defines our emission probabilities (where the model needs to learn the $\phi$ parameters). But the Bayesian learning relies on estimating the posterior probability of the model $\text{P}(\theta,\phi|\mathbf{w})$ (where $\mathbf{w}$ is observed data) with a likelihood $\text{P}(\mathbf{w},\theta,\phi)$ and prior $\text{P}(\theta,\phi)$ terms. So, to implement the Bayesian learning for the HMM, \cite{johnson2007doesn} extends the basic architecture such that the prior probabilities of the parameters can be calculated. 
\begin{nalign}
\label{BayHMM}
    &\theta_y\ |\ \alpha_y\ && \sim\ && \text{Dir}(\alpha_y) \\
    &\phi_y\ |\ \alpha_w\ && \sim\ && \text{Dir}(\alpha_w) \\
    &y_i\ |\ y_{i-1}=y\ && \sim\ && \text{Multi}(\theta_y) \\
    &w_i\ |\ y_{i}=y\ && \sim\ && \text{Multi}(\phi_y)
\end{nalign}
In \ref{BayHMM}, we see that our transition and emission distributions are sampled from Dirichlet distributions, used here because it is a conjugate prior for the multinomial distribution, parameterized by $\alpha_y$ and $\alpha_w$ respectively. These parameters control the sparsity of the probabilities, so as $\alpha_w$ gets closer to zero, the model prefers the hidden states to emit few words \citep{johnson2007doesn}. This feature results in most word types belonging to one pos because of the pressure to minimize non-zero emission probabilities. Ultimately, 
this sparse emission distribution produces the desired peaky tag distribution (figure \ref{fig:JohnsonTagDist}) as most of the tokens will be tagged by a limited number of hidden states. 

Using this Bayesian model specification, \cite{johnson2007doesn} tested two Bayesian estimators with the assumption that the more accurate tag distribution afforded by these estimators will result in improved performance over the vanilla model: Variational Bayes (VB) and Gibbs sampling.
In this work, VB outperforms the vanilla model in one-to-one accuracy but only achieves comparable performance (0.60) to the vanilla model in many-to-one accuracy while Gibbs sampling lags behind on both measures. In a follow-up work, \cite{gao2007comparison} further investigate the performance of these Bayesian estimators by testing the models on varying corpus sizes and varying the Gibbs sampler.
Here, we see the Bayesian estimators outperforming the vanilla model more consistently. In particular, \cite{gao2007comparison} found that Gibbs sampling achieved a noticeable improvement when the corpus size decreased. They hypothesized that because the likelihood term decreases with a smaller corpus, the prior will play a larger role which is evidence for the sparse prior assisting the learning.

Thinking of the implications of these results for language acquisition research, many fundamental debates center around the prior knowledge children may need to achieve their learning goals \citep{chomsky1965, perforstenenbaumregier2011}. While these results are not directly addressing children's preference for a particular prior bias, they suggest that with less evidence assuming a sparse prior that better aligns with the actual label distribution improves tagging performance. 
Additionally, on a practical implementation note, these results impact language acquisition researchers who are increasingly implementing Bayesian models for learning \citep{pearl2020modeling} and often use smaller corpora of child directed speech. 


Besides changing the learning algorithm of the vanilla HMM, we can vary the information that the model is learning from. In this way, \cite{berg2010painless} extended the HMM architecture such that the model could learn from orthographic features.
In particular, the authors alter the emission probability distribution (parameter $\phi$) such that instead of the traditional conditional probability of a word, $w_i$, and its tag, $y_i$, $P(w_i|y_i)$, the emission distribution was the output of a logistic regression: 
\begin{equation} 
    \phi_{w,y}(\mathbf{v}) = \frac{\text{exp}\langle \mathbf{v}, \textbf{f}(w,z)\rangle}{\sum_{w'}\text{exp}\langle \mathbf{v}, \textbf{f}(w',y)\rangle}
\end{equation}
Under this logistic regression, the emission distribution is learned by leveraging some features encoded in our feature function, $\textbf{f}$, with our weight vector, $\mathbf{v}$, which encodes the relative importance of each feature. So, we need to specify the features in $\textbf{f}$, 
and \cite{berg2010painless} selects a few orthographic features. Some of the features are simply indicator variables checking if the word contains a certain character- containing a digit, containing a hyphen, beginning with a capital letter. The remainder are indicator functions checking if the word contains character n-grams up to length 3. While these orthographic features are relatively simple and aren't attempting to make much contact with linguistic theory, the character n-gram features will pick out affixes that words share which are often applied to words in the same syntactic category, 
and more generally this additional orthographic information overlaps with some linguistics work that suggests morphological information may be helpful in learning syntactic categories \citep{clark2003combining}. 

With the model structure specified, \cite{berg2010painless} use the Wall Street Journal (WSJ) corpus (which is written adult-directed text) 
as their data and attempt to fit their log-linear HMM by optimizing the marginal likelihood of the observations. They outline a modification to the EM algorithm suitable for their model framework, but ultimately report best performance with a more direct optimization algorithm (using a gradient based search algorithm L-BFGS).
With this direct optimization, \cite{berg2010painless} reports many-to-one accuracy on the WSJ of 0.755 which was the state of the art performance for unsupervised pos tagging for many years. More modern implementations of this model report further improvement in accuracy (\citealt{tran2016unsupervised} report 0.791 accuracy) by allowing a neural network to learn the weight vector as opposed to hand selecting the relevant orthographic features (transforming the model into a neuralized HMM). 
The initial work from \cite{berg2010painless} and further developments highlight the benefit of providing the model additional information to learn syntactic categories. 

Both \cite{johnson2007doesn} and \cite{berg2010painless} have extended the vanilla HMM which resulted in some modifications to the classic HMM architecture. These modifications have been in service of varying the learning algorithm and providing additional information for the model to learn from, but the resulting output provided emission probabilities and transition probabilities for some corpus data similar to the classic HMM. This next extension will introduce a more fundamental alteration to the HMM architecture where an additional parameter must be learned in addition to the emission and transition probabilities. \cite{stratos2016unsupervised} extended the HMM architecture by adding an ``anchoring" constraint which is defined as follows ($Y$ is the set of tags and $W$ is the set of words or observed states): 
\begin{align}
    A(y) = w : p(w|y)>0 \wedge p(w|y') = 0 \ \forall y' \neq y
\end{align}
With this additional constraint, the model needs to identify an anchor, $w$, for each tag which is a word that the tag emits with nonzero probability \textit{and} no other tag emits this word. Intuitively, this anchoring corresponds to identifying a prototypical word that a learner can build the categories around. Conceptually, this approach shares some motivation to the semantic learning of syntactic categories reviewed in Section \ref{sec:acqgm}. Although this model is not encoding any meaning, the notion of providing the learner the opportunity to ground the categories to some lexical item is shared across the approaches. The semantic approach proposes starting the grammatical learning process by grouping certain meanings, but the learner still anchors the more abstract, complete syntactic categories with a subset of words and their meanings. We will return to discussing the implications of this additional anchoring constraint in section \ref{sec:addinfo}. 

\begin{figure}
    \centering
    \includegraphics[width=\linewidth]{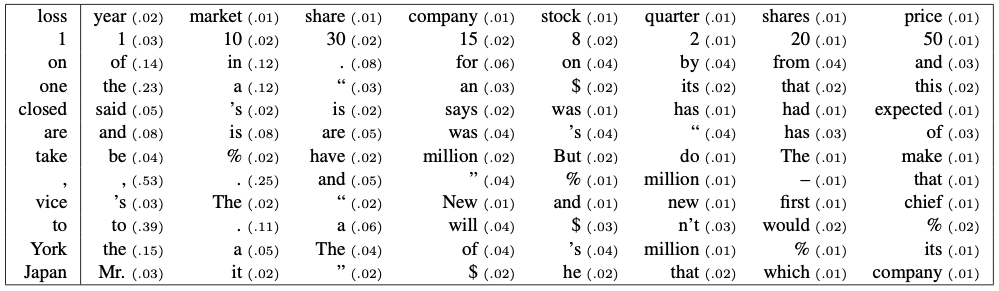}
    \caption{Taken from \cite{stratos2016unsupervised}, we see the learned anchors on the left and the rows list the highest probability emissions for each category.}
\label{fig:anchor}
\end{figure}

Returning to the specifics of the anchor HMM (A-HMM), \cite{stratos2016unsupervised} provide examples of an anchor in English using the gold standard tags (``the" for the determiner category and ``laughed" for the verb category), but the only knowledge the model has about the gold standard categories is the number of categories. 
So, the model must learn the anchors from the data, and then we can match the categories to the gold standard categories by the learned anchors. To illustrate the categories that the model is learning, I included figure \ref{fig:anchor} which lists the anchors and highest probability emissions. The authors ran the model on the universal treebank dataset \citep{mcdonald2013universal} which uses a collapsed set of 12 pos tags which is why we see 12 anchors. 

From figure \ref{fig:anchor}, the authors note that some of the learned categories loosely correspond to the gold standard categories- the ``loss" group seems to be noun-like, the ``on" group seems to be preposition like, ``one" determiner like, and ``closed" verb like. But even these more typical groupings exhibit some interesting behavior as unexpected characters are emitted with relatively high probability: \textbf{.}, \textbf{``}, \textbf{'s}. Additionally, the behavior of the other groupings are a bit difficult to interpret with respect to gold standard groupings, and we see that frequent words like ``the" and ``of" are emitted by multiple labels. 

Because of this abnormality, \cite{stratos2016unsupervised} attempted to provide anchors that corresponded to the gold standard labels in order to better align the categories with the desired groupings. But this manipulation did not improve performance, and \cite{stratos2016unsupervised} explain that this hit in performance compared to a data-driven method is expected with how tied anchor selection is to the model parameter estimation.
Further, \cite{stratos2016unsupervised} remarks that some gold standard tags occur very infrequently, so forcing the model to reserve a hidden state for this tag would decrease the probability of the model. The results from this attempt to more tightly constrain the learned categories seems to illustrate some irreconcilable differences between the data driven learned groupings and the gold standard groupings as reserving a hidden state for very infrequent tags (which the gold standard tags include) is a dispreferred strategy when maximizing the probability of the observed data. This discrepancy offers a place of contact between the engineering approach and the Frequent Frames approach reviewed in section \ref{sec:complearning} where frequency plays a central role in learning categories (FF is implementing a frequency cut-off in the groupings and frequency influences the probability of the models in the engineering approach) while the gold standard labels are agnostic to frequency. 

Turning to performance of the A-HMM (anchored HMM), we first note that \cite{stratos2016unsupervised} also include an augmented A-HMM with the orthographic features from the \cite{berg2010painless} model in the analysis. To situate the A-HMM model in terms of performance, \cite{stratos2016unsupervised} report many-to-one performance on the WSJ with the full set of 45 tags (the most popular evaluation in the unsupervised pos literature). On this task, the augmented A-HMM is competitive with the best performance at the time (the \citealt{berg2010painless} model) achieving 0.677 accuracy compared to the gold standard of 0.749.
The authors explained that learning the full set of 45 tags requires fine grained orthographic knowledge (e.g. categorizing ``played" VBD versus ``play" VBZ). But the effectiveness of this model is evident in its crosslinguistic performance. When running the model on 10 languages from universal treebank dataset using a collapsed set of 12 pos tags, the A-HMM augmented with features outperforms (many-to-one accuracy) baselines including the \cite{berg2010painless} model on the majority of languages. This methodology of testing the model on cross linguistic data with a collapsed tag set is potentially more relevant to acquisition research than the popular WSJ with the full 45 set of tags. 
The full set of 45 tags will have fine grained distinctions that may not be relevant for children first trying to categorize the words in their input. 
For example, returning to our fine grained ``played" vs. ``play" distinction, children may make some tense errors and incorrectly use ``played" while still effectively employing knowledge about a coarser verb-like grouping. 

Overall, the promising performance of the A-HMM illustrates the utility of identifying an anchor word during pos induction-- a strategy that seems effective across many languages. This technique also produces more interpretable category outputs from the model with respect to the gold standard groupings because we can see how the gold standard groupings label the anchors.



\subsection{beyond the HMM}
The HMM is certainly the dominant model structure for unsupervised pos tagging, but other model structures 
have proven successful. One model that has produced competitive results for over 25 years and differs from the standard HMM in interesting ways is the Brown model \citep{brown1992class}. In a review of unsupervised pos tagging models, \cite{christodoulopoulos2010two} report that despite being the oldest and simplest model tested, the Brown model, incredibly, performed competitively. This model remains a strong baseline in the unsupervised pos literature. So, it is worth investigating what makes this model appropriate for this task. 

Interestingly, in the original formulation, \cite{brown1992class} do not set out to learn gold standard pos tags, but they are simply interested in the basic engineering problem of predicting a word given a context of text.
They suggest that adding word classes to a language model will assist in solving this engineering problem and help ``tease from words their linguistic secrets" \citep{brown1992class}. The structure of their class based language model shares the Markov assumption defined in equation \ref{eq:markov} where the probability of a word depends the particular word class and the previous word class. Where it differs from the HMM is how it learns these classes. \cite{brown1992class} showed that learning the optimal grouping of words into classes which maximizes the probability of some training text only depends on the average mutual information between adjacent classes (seen in equation \ref{eq:mutinfo}) as the training size goes to infinity. This objective corresponds to the intuition that we want a labeling scheme for the training text which will make the labels as informative as the following label as possible.

\begin{align} \label{eq:mutinfo}
    I(y_1, y_2) = \sum_{y_1,y_2}\text{p}(y_1y_2) \cdot \text{log}\frac{\text{p}(y_2|y_1)}{\text{p}(y_2)}
\end{align}
This equation also illustrates the simplicity of these computations, one of the advantages of the brown algorithm, as these probabilities are calculated with counts from the training text. Continuing with the learning process, \cite{brown1992class} would want to identify such a partition that maximizes the mutual information, but they note that they are unaware of a practical way to directly identify this partition (a problem that persists today given how many possible ways one could group words into classes). Despite this limitation, they proceed with a greedy clustering algorithm. First, to initialize the Brown model, each word forms its own class.\footnote{For larger text corpora, only the most frequent words form their own category.} Then, the algorithm proceeds by iteratively merging the pair of classes which will result in the least loss of average mutual information between adjacent classes until it reaches a set number of classes. This learning procedure differs from the vanilla HMM because the HMMs require specifying the number of classes and then learning the probabilities associated with these classes as opposed to iteratively clustering based on some objective.
As mentioned earlier, \cite{brown1992class} are not formulating their model for unsupervised pos tagging, but by using simple statistics from the training data (as they rely on counts for computing probability) and a greedy clustering algorithm, the Brown model achieves competitive performance on the standard WSJ many-to-one tagging task (\citealt{christodoulopoulos2010two} report $\sim0.68$). 

In addition to reviewing unsupervised pos tagging models, \cite{christodoulopoulos2010two} extend these models with a prototype based learning model inspired by the model from \cite{haghighi2006prototype}. In particular, they found that the prototype extension to the Brown model produced the best results as \cite{christodoulopoulos2010two} report 76.1 many-to-one accuracy on WSJ which was state of the art at the time of the article and remains competitive. The prototype extension takes the output of a unsupervised model and identifies prototypes for each category which are frequent words that are similar to the other words in the category but dissimilar from the words in other categories (similarity and dissimilarity found by comparing context vectors). These prototypes are fed into the model from \cite{haghighi2006prototype} which form categories around the identified prototypes with a log-linear model with orthographic features. As previously mentioned, \cite{christodoulopoulos2010two} report that the brown model produces the best prototypes among the unsupervised models. 

The final model we will cover which shares our interest in the longevity of the Brown Model is the model proposed by \cite{stratos2018mutual}. Similar to the Brown model, \cite{stratos2018mutual} uses an information theoretic objective to learn pos labels for some text. They first define relevant random variables: $W$ is a random word, $X$ is the context surrounding $W$ (in the best performing model is two previous and two following words), and $Y$ is an unobserved label for $W$.
They wish to maximize the mutual information between $X$ and $Y$, $I(X, Y)$, corresponding to identifying the labels that are most informative about the raw data, 
but this objective is intractable. Instead, they use an established lower bound, called variational lower bound \citep{mcallester2018information}, which minimizes the cross entropy between the conditional label distributions $p(y|x)$ and $q(y|w)$, $H(q,p)$. Intuitively, this objective corresponds to selecting labels that provides as much information as possible about the context and the word-- an objective that nicely follows our understanding of the role that syntactic categories play. Calculating probabilities under the \cite{stratos2018mutual} is more sophisticated than the Brown model as \cite{stratos2018mutual} uses a form of gradient descent to maximize their objective function. Additionally, they are less directly reliant on simple counts from the training as distribution $p$ is defined with word embeddings, a learned representation appropriate for the objective function, which enables learning from the context, 
and distribution $q$ uses a character LSTM which enables learning from character information. On the standard WSJ many-to-one task, this model achieves results comparable to state of the art performance (\citealt{stratos2018mutual} report peak performance of 80.1), and it outperformed baselines on 7 out of 10 languages in the universal treebank dataset \citep{mcdonald2013universal} using many-to-one accuracy.

\section{engineering and acquisition modeling approaches meet} \label{sec:meet}
Now that we have reviewed a wide range of models from both the acquisition modeling side and the engineering side, we will discuss four major themes that will guide our comparison of these approaches. 

\subsection{beyond distributional learning with additional information} \label{sec:addinfo}
\begin{table}[h] 
\begin{center}
\begin{tabular}{|c | c|} 
\hline
  & \textbf{Additional Information}\\ 
 \hline
 \cite{franketal2013cat} & sentence types \\
 \hline
 \cite{gutmanetal2015} & realistic lexicon of word meanings \\
 \hline 
 \cite{johnson2007doesn} & more realistic tag distribution \\
 \hline
 \cite{berg2010painless} & orthographic information \\
 \hline
 \cite{stratos2016unsupervised} & anchoring constraint \\
 \hline 
\end{tabular}
\caption{Summary table of the additional information proposed in various works that extend the distributional learning.}
\end{center}
\end{table}

When comparing the objective for unsupervised pos tagging and modeling the acquisition of syntactic categories, the learning problem is very similar. In both cases, researchers want a learner that infers unobserved labels from language data without assuming knowledge of the specific categories. In all the models presented here, the learner leverages distributional information to infer some structural grouping of the words. One major overlap between the engineering and the acquisition modeling side is the procedure of extending baseline distributional learning to include other sources of information that might assist in the categorization. From the engineering side, we have seen how constraining the tag distributions to better reflect the gold standard distributions extends the vanilla HMM and improves pos tagging performance \citep{johnson2007doesn}. Additionally, from the engineering side extending the vanilla HMM, we have seen additional orthographic information \citep{berg2010painless, stratos2016unsupervised, stratos2018mutual} improve model performance. Intuitively, orthographic information seems essential for learning syntactic categories as many languages apply affixes to certain categories of words. We also saw that including sentence types (motivated by experimental data from acquisition research) improved model performance \citep{franketal2013cat}. 

The last piece of additional information, which has a long history in acquisition literature, was present in both the engineering and acquisition modeling approaches: grounding a category with some lexical item. This proposal of additional information relevant to learning syntactic categories has its roots in the semantic learning hypothesis introduced in section \ref{sec:acqgm}. This proposal suggested that distributional information may not be the primary source of information children rely on when grouping words into syntactic categories. Instead, children may use semantics by grouping words that share similar meanings (physical objects could be grouped in a noun like category with a semantic grouping). Despite some challenges in forming syntactic categories from word meanings (examples presented earlier such as the noun wiggle) and the fact that evidence supports some reliance on distributional information while categorizing \citep{reeder2013shared}, we have seen multiple computational approaches that suggests semantic information may assist distributional learning. From an acquisition modeling perspective, \cite{gutmanetal2015} showed that only assuming a limited vocabulary for the child learner can improve performance in syntactic labels for prosodic phrases. In particular, this semantic bootstrapping process allowed the learner to ground noun and verb categories which allowed for more complete groupings. The FF modeling also runs into a completeness problem, so this prosodic phrase work provides a potential solution in the form of grounding categories with some limited semantic information. Similarly, from the engineering side, we saw how grounding each category to a lexical item could improve upon the standard HMM performance \citep{stratos2016unsupervised}. In this anchored HMM \citep{stratos2016unsupervised}, the model does not add information about lexical meaning, but the lexical items form the basis for the syntactic categories. In particular, the HMM is constrained such that each category must have an anchor word that only this category emits. Additionally, from the engineering side, we saw that extending unsupervised models with a prototype model produced competitive performance to more recent modeling efforts \citep{christodoulopoulos2010two}. Again, this engineering work does not group based on meaning, but it is identifying lexical items most representative of a certain syntactic category. In summary, this last piece of additional information identifies lexical items that are representative for a syntactic category, but the engineering approaches (prototype and anchoring) differs from the acquisition modeling approach by how these lexical items are selected. The engineering side attempts to identify these anchors through distributional information while the acquisition modeling side attempts to leverage some other child knowledge assumed in the literature (namely, a limited lexicon of word meanings) to identify words that serve as the basis for different categories. Depending on the stage of development and the granularity of the syntactic category the child is learning, a combination of lexical meaning and distributional information may be relevant for identifying prototypes. Future modeling work can extend the engineering approach to corpora of child directed speech and investigate the extent to which semantic information assists identifying prototypes (extend beyond only noun and verb categories to see the granularity of categories that semantic information affords).

As a whole, these proposals of additional sources of information leveraged during categorization presents a range of details that must be accounted for when building models of acquiring syntactic categories. The ultimate goal of acquisition modeling work is to approximate as closely as possible the computations that children are using with all the possible information children leverage from their input. To this end, these additional sources of information proposals outline different possibilities along the dimensions of computations and information leveraged from the input. Future work can investigate how children may be sensitive to these sources of information when categorizing and whether children have a preference for one source over another. Even though the acquisition modeling work often motivated their additional sources of information with experimental evidence of children's sensitivity to the particular information source (e.g. prosody, orthographic information, etc.), experimental evidence for the information servicing syntactic categorization in particular is scarce. Experimental paradigms similar to the experiment in \cite{shi2010syntactic} reviewed earlier could assist in identifying the information used during categorization. For example, during a head turn preference experiment, a researcher could compare responses when providing a certain additional piece of information to trials when the additional information is absent (e.g. orthographic information, or phonological information in the case of the head turn paradigm where stimuli is presented auditorily, points to a certain categorization scheme compared to random orthographic information).

\subsection{cognitive plausibility}
Although the learning problem set-up may be very similar for both computational approaches, a major point of departure for the engineering and acquisition modeling approaches is the difference in motivating goals. The acquisition side is motivated by building more realistic models of children's syntactic categorization learning while the engineering side hopes to automatically produce accurate pos tags from a corpus of text. The engineering researchers are not interested in what computations humans make during syntactic category learning. We can categorize this difference in research goals with the notion of cognitive plausibility of the models. Cognitive plausibility is essential to the work of acquisition modeling because we are building models to better understand children's language learning. But developing specific criterion for judging the cognitive plausibility of a model (such that we can objectively decide which of two models is more cognitively plausible) is difficult. A common dimension of cognitive plausibility is the complexity of the learning algorithm where less complex algorithms are preferred when modeling child learning. This dimension is often concerned with memory limitations as many algorithms rely on perfect memory of the input. In section \ref{sec:complearning}, we noted that some early acquisition modeling work \citep{cartwright1997syntactic, redington1998distributional} may not be suitable for a child learning algorithm because of the unbounded memory assumption. Although the engineering models are not intending to mimic child learning, we can extend this cognitive plausibility consideration to these models. Again, the memory limitation consideration plays a role because the EM and Bayesian algorithms (e.g. Gibbs sampling and variational bayes) assume perfect memory as the algorithms make slight changes to the model parameters while fixing the remainder of the data in order to maximize the probability of the model. While intuitively this perfect memory assumption may not be applicable to child learning, some human memory capacity may be unbounded \citep{camina2017neuroanatomical}, so we can not definitively label perfect memory algorithms as implausible. We know that some conscious memory limitations exist, but the limitations of the unconscious memory relevant for the brain's computations are less clear. Analogously, the need to consider a wide range of computational power for child acquisition models and to move beyond a reliance on intuitions for cognitive plausibility is evidenced by the observation that children are sensitive to the statistics of their input \citep{saffranetal1996}. Intuitively, assuming a child has knowledge of the statistics in their input may seem implausible because they do not seem to have conscious awareness of mathematics, but when investigating the computations underlying children's behavior, we must respect the brain's computational abilities that fly under the radar of conscious thought. 

While computational complexity and memory limitations are often central to discussions of a model's cognitive plausibility, it is important to note that there are other dimensions to consider \citep{pearl2020modeling}. Another consideration is the language input that the modeled learner receives. All the engineering models relied on adult text corpora which is not plausible for a child learner. But, the acquisition models provide child-directed speech as the input in order to better approximate the learning environment for children. Relatedly, a cognitively plausible acquisition model will specify the learning period because the particular stage of development is relevant for the language input and the assumed knowledge available to assist the current learning problem. 

As we have seen, cognitively plausibility is central to acquisition modeling. The intuitive belief that simpler models are preferred can help explain the popularity of the FF algorithm \citep{mintz2003} as it is the simplest learning algorithm we reviewed. FF only requires the learner to remember the words that appear in frequently occurring frames using simple counts from the input. Further work even illustrated how this learning could be done incrementally \citep{wangmintz2008} in a more plausible setting. But acquisition modeling work should not ignore more complex learning algorithms based on intuitive notions of cognitive plausibility. Future work formulating more precise considerations for cognitive complexity can assist in judging the plausibility of various computations in service of building plausible acquisition models that incorporate plausible computations on realistic input.

\subsection{context}

\begin{table}[h] 
\begin{center}
\begin{tabular}{|c | c | c | c | c|} 
\hline
  & Frequent Frames & HMM & Brown & \cite{stratos2018mutual}\\ 
 \hline
 \textbf{Relevant Context} & \begin{tabular}{@{}c@{}}previous and following \\ lexical item \end{tabular} & previous tag & previous tag & \begin{tabular}{@{}c@{}}two previous and two \\ following lexical items \end{tabular}\\
 \hline
\end{tabular}
\caption{Summary table of the relevant context for categorizing a target word under the main model architectures.}
\end{center}
\end{table}

Another theme relevant to all the reviewed models is the specification of context that the learner leverages while categorizing. How many surrounding words are relevant for labeling a certain word? Are preceding words or succeeding words considered? How may a context of surrounding syntactic category labels assist learning? 

Starting with FF \citep{mintz2003}, the relevant context in determining the syntactic category label for a certain word is the preceding and following lexical item. \cite{chemlaetal2009} investigated why this specific context seems to produce good results. They found that frame structure does produce the best performance as compared to other two-word contexts (two preceding words and two following words). The authors suggest that the frame structure better constrains the syntactic category of the target word. They also attempted to create more complete FF categories by recursively applying the FF algorithm. This involves applying the output of a FF grouping to the text such that the context could be category labels or lexical items. For example, if ``I" and ``you" are grouped together with the first FF application, then the two frames [I X it] and [you X it] could now group the target words together with the [LABEL X it] frame \citep{chemlaetal2009}. With this manipulation, the relevant context is preceding and following lexical items \textit{or} their syntactic category label. But, this recursive FF application hurt performance encouraging \cite{chemlaetal2009} to conclude that the strength of FF lies in the groupings of surrounding lexical items (as they speculate that this frame structure produces more stable environments for syntactic grouping).

Turning to the context specified by other models, the relevant context for the dominant engineering architecture, HMM, is the previous tag and the current tag. Because the HMM is a generative model, discussing the relevant context is less straightforward. With FF, we thought about how surrounding lexical items informs the syntactic category of a target word, but the generative process of HMM assumes that a tag is generated using the transition probabilities of the previous tag, and then the tag generates a certain word using the emission probabilities for that tag. 
But because the relevant probabilities for tagging a word is emission probabilities for that word and the transition probabilities from the previous tag, we can consider these pieces as the relevant context under the HMM structure. There is variation in the relevant context for HMM models as the initial Bayesian HMM formulation in \cite{goldwatergriffiths2007}, the model present in the acquisition modeling work of \cite{franketal2013cat}, uses a trigram HMM structure. Here, the transition probabilities depend not on the previous tag, but on the previous two tags. Overall, we can contrast the HMM context with the FF context in that it includes previous labels as relevant context and no succeeding information is relevant due to the generative process. 

As previously discussed, the Brown model \citep{brown1992class} has a similar structure to the HMM models. The relevant statistic for the Brown model's clustering procedure is the mutual information between adjacent tags, so we can again think of the relevant context for categorizing a target word as the previous tag. Finally, in \cite{stratos2018mutual} which was inspired by the Brown model, they manipulated the context window but ultimately found that best performance was achieved with a window of two previous and two following lexical items. The FF model and this model from \cite{stratos2018mutual} are the only models reviewed where the relevant context for distributional learning includes lexical items. While \cite{stratos2018mutual} is attempting to identify the labeling scheme that minimize the cross entropy between the context conditional distribution, $p$, and the lexical item conditional distribution, $q$, FF seems to be creating labels which are informative about the one word context frame without learning from the lexical item. Another difference between the two algorithms is \cite{stratos2018mutual} pre-specifies the number of labels used to categorize all the words in the corpus while FF creates some number of labels based on the particular frequency cut-off used and does not produce full coverage of the corpora. If we maintain that the FF structure is relevant to children's acquisition of syntactic categories \citep{mintz2006}, a potential update to this structure could alter the model from \cite{stratos2018mutual} to optimize the context conditional distribution, $p$, and define the context in the same way as FF (a window of lexical items size 1). This alteration would eliminate the learning from the orthographic information in the lexical item and utilize the FF contexts to more closely mimic the original FF model. This extension would also investigate the limits of the FF architecture because it produces full coverage for a corpus of text relying only on the distributional information present in the FF context window. 

Overall, we have seen how context may vary with size and vary with context of lexical items versus syntactic category labels. Future computational work may vary the lexical vs. label context to better tease apart the advantages of each approach, but we have seen, at least under the FF architecture, context of lexical items seems to be important for syntactic categorization. Similar to the future work proposed for further investigating the additional information, future experimental work could manipulate the size of the lexical context when children perform categorization tasks.

\subsection{evaluation metrics} \label{sec:evals}
All of these modeling approaches need to confront the question of evaluation. How can we judge the success of a model? In the unsupervised setting, where categories are learned without knowledge of the gold standard tags, this evaluation is particularly difficult because researchers first need to map the model's groupings onto the gold standard labels. We saw how various metrics achieved this feat (with many-to-one accuracy being the most popular, but one-to-one accuracy is also common). We also saw how information theoretic measures avoided the mapping problem and directly compared the model output as a distribution to the gold-standard distribution. Additionally, we saw how evaluation could be divided into accuracy and completeness to assess different dimensions of the model performance. While investigating differences in these metrics and looking for ways to make improvements may be fruitful avenues of research, the observation about all these evaluation metrics most relevant for acquisition research is that they all rely on gold standard tags. 

While a gold standard evaluation may be appropriate for the engineering approach where the goal is to reproduce these gold standard tags, this evaluation seems to be less appropriate for the acquisition modeling setting. The goal of the acquisition modeling approach is to approximate how children learn syntactic categories. So, evaluating model performance with gold standard metrics may give us insight into potential computations for child learning if we assume that children's syntactic groupings are similar to the gold standard groupings. Under this assumption, improving model performance as evaluated by a gold standard metric could also represent an improvement in the closeness of approximation to children's computations. But, this assumption will certainly be too strong especially in the case of a full 45 English tag set where many fine-grained distinctions are encoded. A more reasonable perspective for evaluating acquisition models with gold standard metrics would assume that the adult target knowledge that children are acquiring is the gold standard groupings, so the models are proposing potential computations that children may use to arrive at the adult representation. Investigating the validity of this assumption may be difficult, but a better understanding of adult syntactic category representation and how closely it aligns with our gold standard groupings would greatly assist in determining how gold standard evaluations align with the goals of modeling the acquisition of syntactic categories. 

Overall, the reliance of the gold standard evaluations pulls focus away from the question of why children would want to acquire these syntactic categories in the first place. If we are treating these gold standard labels as an established fact of language relevant for language processing then it follows that children must acquire them. But when considering there are discrepancies in linguists' definitions of parts of speech and variation exists across languages, the question of why remains open. \cite{barseverpearl2016} proposes an answer to this question and offers an alternative evaluation metric that does not rely on gold standard tags. The main intuition behind this work is that an appropriate labeling scheme should ease processing for future input. So, children have an incentive to learn syntactic categories because they will be able to process their language input more easily. While many of the models reviewed in this paper are optimizing the probability of the data (and therefore identifying labels that are high probability and easier to process), a fruitful future direction would focus on the language processing benefit and investigate how different models could reduce processing difficulty for future input.

\subsection{conclusion}
In this work, we set out to investigate the potential computations children rely on while acquiring syntactic categories. Although syntactic categories may be difficult to define, we saw their usefulness in generalizing words to new context, and we saw how syntactic categories play a central role in linguistic productivity. We then motivated our assumption that children are learning some syntactic groupings with experimental evidence before turning to various proposals for the potential computations underlying this learning process. The bulk of this work was devoted to reviewing various acquisition models and engineering models (where the unsupervised pos tagging set-up closely mimicked the learning problem outlined by the acquisition approach) in order to better understand the range of possibilities for the computations underlying syntactic category learning. Comparing these approaches illuminated different dimensions (relevant context and additional information beyond distributional information) along which these potential computations could vary. 

In acquisition modeling, we are often interested in investigating the computations performed by children when learning language, but we often focus on investigations of ``potential" computations without the ability to directly observe computations. This problem is particularly acute for modeling syntactic category acquisition where the output is a representation of word groupings that is difficult to probe experimentally. Additionally, in light of the previous evaluation discussion, there is not much consensus on the target representation children want to acquire. All these outstanding questions offer exciting opportunities to advance our understanding. Through experimental work, we can better understand how these various model assumptions could correspond to children's behavior (investigating how children could leverage varying contexts and information beyond distributional information during syntactic categorization). Future work can also apply various engineering models to more realistic child learning settings and develop more principled evaluation metrics appropriate for the child's learning goals. Finally, future work can situate the acquisition of syntactic categories within other categorization tasks that children may perform (e.g. learning phonological classes may also be learned from distributional information; \citealt{mayer2020algorithm}). Thinking about the similarity of this computational task to related tasks could assist in developing more general theories about the mental computations children employ.

\bibliographystyle{apalike}
\bibliography{latexrefs}

\end{document}